\pdfoutput=1

\documentclass[sigconf,nonacm]{acmart}
\makeatletter\@ACM@balancefalse\makeatother
\usepackage{flushend}   %% balances the last page automatically
\usepackage{balance}    %% \balance is used once, on the page before the landscape figure
\makeatletter\def\@titlefont{\fontsize{15}{19}\selectfont\sffamily\bfseries}\makeatother

\usepackage{graphicx}
\usepackage{capt-of}
\usepackage{etoolbox}
\usepackage{stfloats}   %% allows [b] placement for figure*
\patchcmd{\maketitle}{\setcounter{footnote}{1}}{\setcounter{footnote}{0}}{}{}
\patchcmd{\maketitle}{\setcounter{footnote}{1}}{\setcounter{footnote}{0}}{}{}
\usepackage{booktabs}
\usepackage{enumitem}
\graphicspath{{figures/}}

\copyrightyear{2026}
\acmYear{2026}
\setcopyright{none}
\acmConference{}{}{}
\acmDOI{}
\acmISBN{}
\renewcommand\footnotetextcopyrightpermission[1]{}

\begin{document}
%% Allow single lines at column breaks so columns fill completely
\clubpenalty=150 \widowpenalty=150

\title{ANASSA: An Agentic AI Orchestration Framework for Spatial Intelligence}

\author{Constantinos Papantoniou}
\authornote{Corresponding author: Constantinos Papantoniou $\cdot$ contact@anassa.ai. LinkedIn: \url{https://www.linkedin.com/in/constantinos-papantoniou/}}
\affiliation{%
  \institution{Anassa.ai}
  \city{}%% TODO: add city
  \country{USA}}

\author{Brian Hilton}
\authornote{LinkedIn: \url{https://www.linkedin.com/in/brian-hilton-ph-d/}}
\affiliation{%
  \institution{Research Scientist}
  \city{}%% TODO: add city
  \country{USA}}

\renewcommand{\shortauthors}{Papantoniou and Hilton}

\begin{abstract}
The emergence of large language models (LLMs) and large multimodal models (LMMs) has enabled a new class of agentic systems capable of integrating natural language understanding with tool-based execution. In geographic information systems (GIS), this shift is transforming traditional, expert-driven workflows into semi-autonomous systems that can interpret user intent, construct spatial workflows, and execute geospatial analysis tasks. However, existing approaches remain limited by fragmented integration of reasoning, execution, and evaluation, particularly in complex, real-world environments.

This study synthesizes recent advances in agentic GIS frameworks, benchmarks, and surveys to identify limitations in spatial reasoning, execution robustness, validation, governance, and evaluation. Building on these insights, it introduces ANASSA (Autonomous Neural Agents for Spatial Systems Architecture), an agentic AI orchestration framework that integrates structured spatial reasoning, multi-agent workflow orchestration, execution feedback, authoritative spatial validation, provenance, uncertainty handling, and human decision authority within a unified system design. The contribution is an architecture-level specification: eleven components across four layers, a six-step Geospatial AI Cognitive Loop, cross-component contracts, and governance mechanisms intended to make agentic geospatial workflows traceable, reproducible, and accountable. Empirical performance evaluation is reserved for implementation and deployment studies.
\end{abstract}




\maketitle

%% ===================================================================
\section{Introduction}

Modern geospatial operations face persistent coordination challenges across organizational boundaries. In governments, enterprises, defense agencies, and research institutions, spatial data is distributed through disconnected systems, including GIS platforms, satellite feeds, IoT sensors, field devices, cloud databases, and legacy infrastructure, each operating largely in isolation~\cite{papantoniou2021enterprise}. Human analysts are required to manually extract, transform, route, and reconcile data across these fragmented systems, creating coordination overhead that diverts cognitive bandwidth from spatial reasoning and decision support~\cite{kaza2026tool,mansourian2026bridging}. Recent advances in agentic AI suggest that such coordination tasks can increasingly be delegated to autonomous systems capable of planning, memory, and tool orchestration~\cite{wang2024survey,xi2023rise}.

The integration of large language models (LLMs) and large multimodal models (LMMs), such as GPT, Claude, and Gemini, into geographic information systems (GIS) marks a transition from traditional tool-based analysis toward interactive and semi-autonomous spatial intelligence systems. Conventional GIS workflows require domain expertise and manual specification of spatial operations, whereas emerging agentic approaches enable natural language interaction with geospatial data by combining reasoning, retrieval, and tool invocation~\cite{kaza2026tool,liang2026geoagenticrag}. This shift is further contextualized by broader developments in agentic artificial intelligence (AI), where systems embed planning, memory, and tool orchestration capabilities to perceive, reason, and act within complex environments~\cite{hashemi2026survey,talemi2026agentic}.

A key methodological advancement is the emergence of multi-agent architectures integrated with retrieval-augmented generation (RAG). Traditional RAG systems are limited in geospatial contexts because they rely on text similarity and cannot represent spatial semantics such as topology, proximity, and spatial dependencies~\cite{liang2026geoagenticrag}. To address this limitation, recent frameworks such as GeoAgentic-RAG~\cite{liang2026geoagenticrag}, GeoAgent~\cite{hu2025geoagent}, and multi-agent autonomous GIS systems~\cite{mansourian2026bridging} utilize coordinated agents that collaboratively perform task decomposition, data retrieval, spatial analysis, and workflow execution. These systems typically rely on role differentiation (e.g., retrieval agents, reasoning agents, analysis agents) and structured orchestration mechanisms, such as directed acyclic graphs (DAGs), to enable coherent multi-step spatial reasoning~\cite{liang2026geoagenticrag,mansourian2026bridging}. This architectural paradigm aligns with broader research demonstrating that multi-agent systems improve modularity and scalability in complex geospatial workflows~\cite{hashemi2026survey,wu2025democratizing}.

Parallel to multi-agent approaches, recent work highlights the importance of structured workflow representations for aligning natural language queries with geospatial computation. Spatial-Agent introduces a concept transformation framework in which queries are converted into GeoFlow graphs that explicitly encode spatial entities, functional roles, and transformation dependencies~\cite{bao2026spatialagent}. Similarly, GeoFlow formalizes workflow automation using Activity-on-Vertex graphs, in which tasks are decomposed into executable steps with explicit tool-calling objectives~\cite{bhattaram2025geoflow}. These approaches emphasize the need for intermediate representations that constrain reasoning and ensure consistency with GIScience principles. Other frameworks, such as LLM-driven multi-agent GIS pipelines~\cite{mansourian2026bridging}, reinforce this trend by demonstrating how structured reasoning and workflow abstraction improve the translation of natural language into executable spatial processes.

Empirical evidence indicates that these approaches substantially improve performance compared to traditional LLM and RAG baselines. GeoAgentic-RAG reports execution success rates exceeding 85\% and high answer correctness, demonstrating the effectiveness of integrating retrieval, reasoning, and tool execution~\cite{liang2026geoagenticrag}. Benchmarking studies such as GeoBenchX further confirm that LLM-based agents can perform multi-step geospatial reasoning but reveal significant variability depending on task complexity and reasoning depth~\cite{krechetova2025geobenchx}. Similarly, GeoAgentBench demonstrates that performance depends heavily on correct tool sequencing, parameter configuration, and the ability to adapt workflows based on runtime feedback~\cite{yu2026geoagentbench}. These findings indicate that agentic GIS systems perform well in simple and moderately complex workflows but remain challenged by multi-step tasks requiring precise spatial reasoning and parameter inference.

The effectiveness of these systems is also strongly influenced by reasoning paradigms and execution strategies. Reactive strategies, such as ReAct-style agents, improve adaptability by incorporating feedback loops, whereas planning-based approaches offer more structured reasoning but may fail under dynamic conditions~\cite{yu2026geoagentbench}. Hybrid approaches that combine planning and reactive execution such as Plan-and-React frameworks demonstrate improved robustness in long-chain workflows~\cite{yu2026geoagentbench}. Research on multi-granularity and spatio-temporal agent systems further suggests that hierarchical reasoning and distributed collaboration can enhance scalability and performance across heterogeneous geospatial tasks~\cite{hashemi2026survey,wu2025democratizing}. These findings are consistent with broader perspectives that emphasize the evolving role of GIS systems as collaborative agents rather than static analytical tools~\cite{kaza2026tool}.

Despite these advances, several persistent limitations are identified in the literature. First, spatial reasoning remains incomplete, particularly for quantitative analysis and ambiguous spatial predicates, where systems may produce outputs that are syntactically valid but contextually incorrect~\cite{dorobantu2026geospatial,liang2026geoagenticrag}. Second, execution robustness remains a major challenge, as real-world GIS workflows involve heterogeneous data, coordinate transformations, and topological constraints that are difficult for agents to fully manage autonomously~\cite{hu2025geoagent,yu2026geoagentbench}. Third, performance is highly dependent on data quality and metadata, limiting generalization to environments with incomplete or poorly structured datasets~\cite{liang2026geoagenticrag,mansourian2026bridging}.

Additional challenges emerge in the context of multimodal geospatial data, particularly in remote sensing applications. Agentic AI systems integrating satellite imagery, LiDAR, and other modalities face limitations in long-horizon reasoning, memory management, and cross-modal integration~\cite{hashemi2026survey,talemi2026agentic}. These constraints highlight the difficulty of scaling current approaches to real-world geospatial intelligence tasks involving temporal and multi-source complexity.

Finally, evaluation frameworks remain limited. Many benchmarks are static or simplified and do not capture real-world complexity. Although dynamic frameworks such as GeoAgentBench~\cite{yu2026geoagentbench} and GeoBenchX~\cite{krechetova2025geobenchx} advance execution-based and multi-step assessment, standardized protocols for multimodal data, temporal dynamics, and long-chain reasoning are still lacking~\cite{dorobantu2026geospatial,krechetova2025geobenchx}. Together with the gaps in spatial reasoning and execution robustness identified above, these limitations characterize current agentic GIS systems as semi-autonomous analytical agents rather than fully grounded geospatial intelligence systems.

This study introduces ANASSA (Autonomous Neural Agents for Spatial Systems Architecture), an orchestration architecture for geospatial intelligence; \emph{neural} denotes the foundation-model (LLM/LMM) substrate on which its agents operate. Building on prior research~\cite{papantoniou2021enterprise}, ANASSA extends the GeoBlockchain coordination model~\cite{papantoniou2021geoblockchain} from blockchain-based coordination to AI-agent-based orchestration, distributing reasoning, planning, validation, execution, and learning across specialized agents under explicit governance. The framework addresses the identified integration gap by coupling spatial reasoning with execution-aware orchestration, feedback-driven adaptation, authoritative spatial evidence, provenance, uncertainty handling, and human decision authority.

ANASSA does not claim novelty for multi-agent orchestration, human-in-the-loop approval, feedback loops, or authoritative-data validation in isolation. Its contribution is the spatially grounded integration of these mechanisms within one orchestration architecture, with explicit cross-component contracts and validation pathways that connect spatial reasoning, geospatial execution, governance, provenance, and learning. This distinction separates the framework-level contribution from established techniques that ANASSA composes and specializes for spatial intelligence.

This paper makes four contributions: (1) it specifies ANASSA as a unified architecture of eleven components across four layers; (2) it defines a six-step Geospatial AI Cognitive Loop that coordinates sensing, reasoning, planning, decision and hallucination validation, execution, and learning; (3) it introduces architecture-level mechanisms for authoritative spatial evidence, data quality, uncertainty, provenance, governance, and human decision authority; and (4) it defines design requirements for component interfaces, execution controls, scalability, and reproducibility that can be evaluated in subsequent implementations.

%% ===================================================================
\section{The ANASSA Framework}

While recent research demonstrates that agentic GIS systems improve automation, workflow generation, and natural language interaction~\cite{hu2025geoagent,liang2026geoagenticrag,mansourian2026bridging}, a consistent gap emerges across the literature: current frameworks lack unified mechanisms for robust spatial reasoning, execution reliability, and evaluation under real-world complexity. Multi-agent architectures improve modular reasoning and task decomposition~\cite{liang2026geoagenticrag,wu2025democratizing}, yet they remain sensitive to ambiguous spatial constraints, implicit parameter selection, and multi-step dependencies~\cite{dorobantu2026geospatial,krechetova2025geobenchx,yu2026geoagentbench}. Similarly, structured reasoning approaches such as GeoFlow graphs and concept transformations enforce logical consistency but often do not fully integrate dynamic execution feedback or multimodal data validation~\cite{bao2026spatialagent,bhattaram2025geoflow}. Benchmarking efforts partially address these limitations but reveal that performance drops in long-chain workflows and heterogeneous data environments, indicating that current models struggle with operational robustness rather than high-level workflow formulation~\cite{krechetova2025geobenchx,yu2026geoagentbench}.

Therefore, the central research gap lies in the integration of three dimensions that remain fragmented across existing research:
\begin{enumerate}[label=\textbf{(\arabic*)},leftmargin=*]
  \item spatially grounded reasoning,
  \item execution-aware workflow orchestration, and
  \item evaluation readiness for realistic, multimodal settings.
\end{enumerate}

Addressing this gap requires moving beyond current semi-autonomous systems toward fully grounded, feedback-driven, and context-aware geospatial agents, where reasoning, execution, and evaluation are co-designed rather than independently optimized. This paper addresses dimensions (1) and (2) directly and prepares for (3) by defining the measurable architectural variables that later evaluation must report.

\begin{figure*}[!b]
  \centering
  \Description{Three horizontal process flows comparing traditional GIS (user, manual query, tool execution, result), LLM/LMM-augmented GIS (user, prompt, LLM/LMM, script, execution, result), and agentic GIS (user, intent, workflow, multi-agent execution, feedback, validated output).}
  \includegraphics[width=\textwidth]{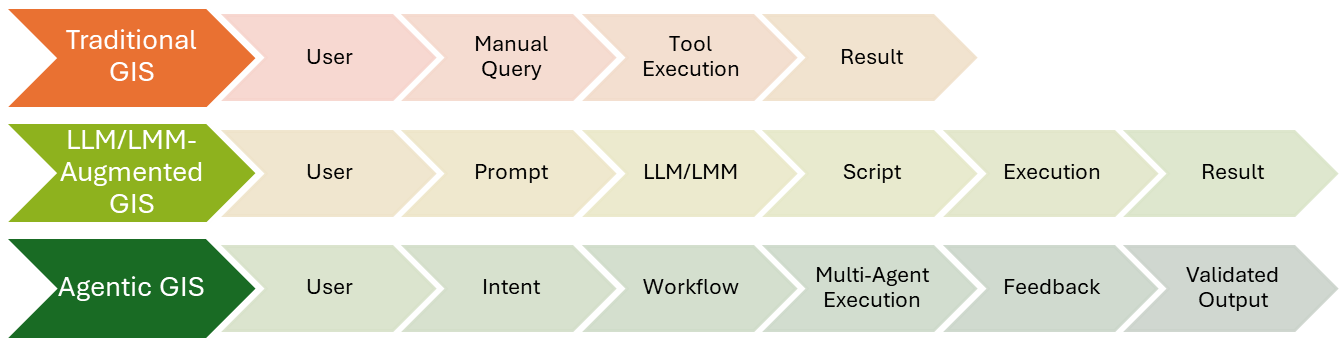}
  \caption{Comparison of GIS workflows from manual processes to agentic GIS with integrated reasoning-execution-evaluation loops.}
  \label{fig:workflows}
\end{figure*}

As a result, this study addresses this gap by specifying a spatial agentic AI framework that integrates reasoning, execution, validation, governance, and feedback within a single system architecture. By emphasizing explicit interfaces, authoritative spatial evidence, runtime controls, and traceable feedback, the framework provides a foundation for implementation and evaluation of robust and scalable agentic GIS systems. This integrated architecture combines natural language understanding, structured workflow generation, multi-agent coordination, validation, and feedback-driven execution. Figure~\ref{fig:workflows} contrasts the resulting agentic GIS workflow with traditional and LLM/LMM-augmented GIS workflows.

The agentic GIS workflow in Figure~\ref{fig:workflows} integrates the three dimensions:
\begin{itemize}[leftmargin=*]
  \item \textbf{Reasoning (User, Intent, Workflow):} Interprets natural language queries and constructs structured workflows,
  \item \textbf{Execution (Multi-Agent Execution):} Invokes geospatial tools and performs analysis,
  \item \textbf{Evaluation (Feedback, Validated Output):} Enables dynamic adaptation through error detection and workflow refinement.
\end{itemize}

Unlike prior approaches that treat reasoning and execution as separate processes, this framework emphasizes feedback-driven integration, enabling continuous interaction between planning and execution stages.

\subsection{Architecture: Eleven Components Across Four Layers}

\begin{figure*}[t]
  \centering
  \Description{Table listing the four ANASSA layers, the components assigned to each layer, and the function of each layer.}
  \includegraphics[width=\textwidth]{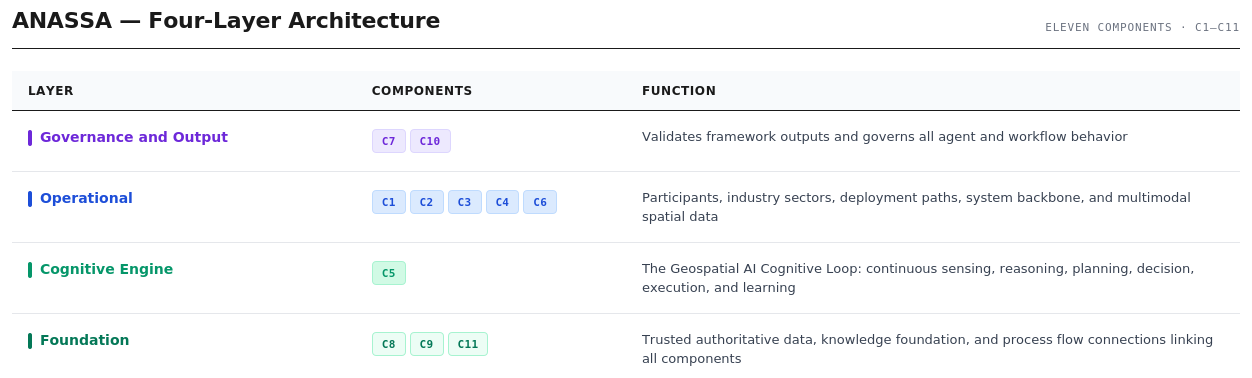}
  \caption{ANASSA four-layer architecture. The eleven components are organized into Governance and Output (C7, C10), Operational (C1--C4, C6), Cognitive Engine (C5), and Foundation (C8, C9, C11).}
  \label{fig:fourlayer}
\end{figure*}

ANASSA is organized into eleven numbered architectural components (C1--C11) structured across four layers (Figures~\ref{fig:fourlayer} and~\ref{fig:highlevel}); Table~\ref{tab:components} summarizes them. The Governance and Output layer (C7, C10) validates framework outputs and governs agent and workflow behavior. The Operational layer (C1--C4, C6) defines participants, industry sectors, deployment paths, the technical system-of-systems backbone, and multimodal spatial data. The Cognitive Engine (C5) runs the six-step Geospatial AI Cognitive Loop. The Foundation view groups C8 with process components C9 and C11 in Figure~\ref{fig:fourlayer}, while Figure~\ref{fig:highlevel} depicts C9 and C11 as vertical cross-layer process flows. Accordingly, C8 provides the authoritative data and knowledge foundation, and C9/C11 are Foundation-associated process components whose flows operate across the architecture.

\begin{table}[!h]
  \caption{The eleven ANASSA components, as depicted in Figures~\ref{fig:fourlayer} and~\ref{fig:highlevel}.}
  \label{tab:components}
  \footnotesize
  \begin{tabular}{@{}p{0.5cm}p{2.4cm}p{5.2cm}@{}}
    \toprule
    \textbf{ID} & \textbf{Component} & \textbf{Role} \\
    \midrule
    C1 & Geospatial Participants & Individuals, teams, departments, organizations, and communities whose intent enters the framework. \\
    C2 & Industry Sectors & Earth and space application domains served, each with sector-specific parameters. \\
    C3 & Deployment Paths & Enterprise, cloud, edge, SaaS, desktop, field, and mobile execution environments. \\
    C4 & Distributed Geospatial System-of-Systems & The ANASSA Orchestrator (Layer 0) and AI infrastructure layers L1--L4. \\
    C5 & Geospatial AI Cognitive Loop & Six-step sensing, reasoning, planning, decision and validation, execution, and learning cycle (Section~\ref{sec:loop}). \\
    C6 & Spatial Data & Multimodal inputs: unstructured, tabular, LiDAR and point cloud, imagery and raster, vector, 3D and CAD/BIM, and real-time IoT streams. \\
    C7 & Governance & Framework-wide overlay of seven governance pillars (Section~\ref{sec:governance}). \\
    C8 & Authoritative Geospatial Data and Knowledge Foundation & Certified repositories, knowledge base, trust and certification, distributed infrastructure, and security and sovereignty. \\
    C9 & Data and Intent Process Flow & Downward flow carrying participant intent and data into the Cognitive Loop. \\
    C10 & Output and Intelligence Delivery & Trusted geospatial intelligence, auditable reasoning chains, transparent decision records, and hallucination-validated outputs. \\
    C11 & Learning and Output Process Flow & Upward flow returning outputs and learning signals to all components. \\
    \bottomrule
  \end{tabular}
\end{table}

Each component is implemented as a coordinated set of specialized agents under the ANASSA Orchestrator (C4). C4 resides in the Operational layer and maintains cross-layer orchestration interfaces. The ``Layer 0'' notation shown within Figure~\ref{fig:highlevel} refers to the Orchestrator's internal infrastructure tier and does not define a fifth ANASSA architectural layer; beneath it, C4 organizes four AI infrastructure layers: a multi-model language gateway (L1), specialized geospatial agents for analysis, cartography, remote sensing, and workflow orchestration (L2), GIS platform and protocol integration (L3), and compute, retrieval, and model-hosting infrastructure (L4). C4 decomposes participant intent into agent task graphs, sequences execution, applies execution controls, monitors intermediate states, and synthesizes validated outputs. After each cycle, the Learning Agent may update permitted participant preference models, industry parameters, agent capabilities, and Orchestrator coordination rules, subject to policy and provenance controls; changes to the governance rules in C7 themselves are proposed by the Learning Agent but require human authorization.

\begin{figure*}[t]
  \centering
  \Description{Table of the six steps of the C5 Geospatial AI Cognitive Loop, S1 to S6, showing the agent responsible for each step, its function, and whether a human-in-the-loop gate applies (yes at S3 and S4).}
  \includegraphics[width=0.9\textwidth]{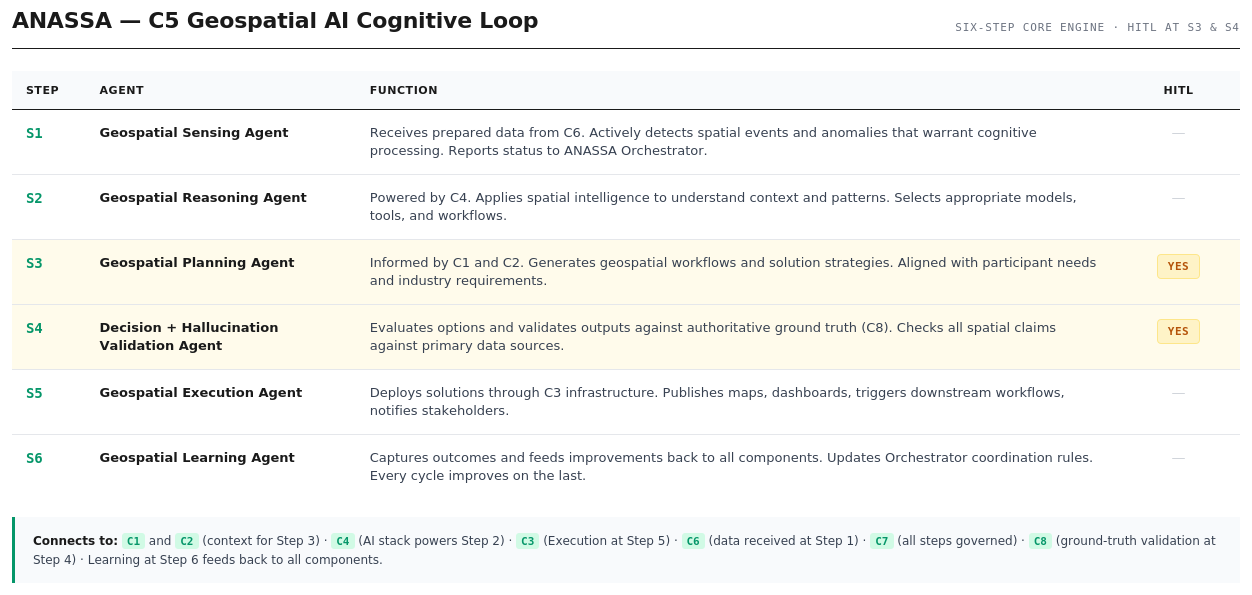}
  \caption{The C5 Geospatial AI Cognitive Loop. Six steps coordinated by the ANASSA Orchestrator, with Human-in-the-Loop checkpoints at Steps S3 (Planning) and S4 (Decision and Hallucination Validation).}
  \label{fig:loop}
\end{figure*}

\subsection{The Geospatial AI Cognitive Loop}
\label{sec:loop}

Component C5 is the heart of ANASSA: a continuous six-step reasoning and action cycle that transforms spatial data into spatial intelligence and intelligence into controlled action (Figure~\ref{fig:loop}). The six steps---Sensing (S1), Reasoning (S2), Planning (S3), Decision and Hallucination Validation (S4), Execution (S5), and Learning (S6)---are each carried out by a dedicated geospatial agent, coordinated by the ANASSA Orchestrator (C4), and grounded in authoritative data from C8. As Figure~\ref{fig:loop} shows, S1 receives prepared data from C6, S2 is powered by the C4 AI stack, S3 draws context from C1 and C2, S4 validates against C8, S5 deploys through C3 infrastructure, and all steps are governed by C7. At S4, hallucination validation is treated as part of a broader spatial-validation function that also evaluates spatial reference, topology, positional accuracy, resolution, temporal validity, uncertainty, jurisdiction, and source authority.

The cycle is feedback-driven rather than linear. Outputs from S5 do not terminate the workflow; instead, they are captured by S6 and fed back into every component, enabling participant preference models, industry parameters, agent capabilities, and Orchestrator coordination rules to adapt over time. The loop is therefore continuous and self-improving within the bounds set by C7. Execution at S5 produces information products and workflow actions---maps, dashboards, downstream workflow triggers, and stakeholder notifications---rather than direct actuation of physical systems, which is outside the scope of this architecture. This closed-loop design distinguishes ANASSA from pipeline-based agentic GIS systems, which typically treat execution as a terminal stage and rely on offline retraining for adaptation~\cite{liang2026geoagenticrag,yu2026geoagentbench}.

Governance checkpoints appear at S3 (Planning) and S4 (Decision and Hallucination Validation). The requirement for explicit human approval is risk- and policy-dependent rather than universally mandatory. High-consequence workflows require human authorization before execution or release; lower-risk workflows may proceed within pre-approved policy bounds when validation criteria are satisfied and the decision path is logged. This preserves explicit human accountability where consequences justify intervention without making human review a fixed latency requirement for every machine-paced operation.

Human expertise enters ANASSA through explicit policies, domain constraints, thresholds, workflow approvals, overrides, correction signals, and adjudication records. The framework distinguishes codifiable knowledge---such as rules, standard operating procedures, and validation thresholds---from expert judgment that depends on context, trade-offs, and institutional responsibility. ANASSA therefore does not assume that decision-maker competence can be completely encoded; instead, it preserves human authority for ambiguous, conflicting, or consequential decisions and records those interventions as part of workflow provenance~\cite{amershi2019guidelines}.

\subsection{Governance, Spatial Validation, and Hallucination Management}
\label{sec:governance}

Component C7 functions as the governance backbone of ANASSA and is positioned as a framework-wide overlay that connects to every other component. Its design is informed by the CPMAI (Cognitive Project Management for AI) methodology, which structures the AI project lifecycle from problem definition through production operation~\cite{cognilytica2019cpmai}. Governance is organized across seven pillars (Figure~\ref{fig:highlevel}): business and problem governance (G1), data governance and sovereignty (G2), AI model and hallucination governance (G3), human-in-the-loop decision authority (G4), trust and auditability (G5), ethical and responsible AI (G6), and operationalization (G7).

A distinctive feature of the framework is the explicit treatment of unsupported spatial claims as a governance and validation concern. Fabricated coordinates, invented boundaries, incorrect transformations, stale observations, topological inconsistencies, and unsupported spatial relationships can produce outputs that appear plausible while being operationally invalid. C7 therefore coordinates claim validation against authoritative spatial evidence maintained through C8, which serves as the framework's single point of authority: agents validate against C8 rather than against arbitrary external sources. Within C8, however, ground truth is a governed determination rather than an assumption of infallibility. C8 maintains multiple certified repositories with trust scores and provenance, and each source is evaluated with respect to authority, provenance, quality, temporal validity, spatial resolution, jurisdiction, and uncertainty. When authoritative sources conflict or evidence is insufficient, the workflow records the conflict and escalates according to governance policy rather than forcing a binary ground-truth decision.

\subsection{Spatial Validation and Uncertainty}

ANASSA treats spatial validity as an explicit structured contract evaluated at S4 rather than as a generic hallucination check. The contract is aligned with established geographic data-quality principles, including the need to describe and evaluate data quality for intended use~\cite{iso19157}.

\begin{itemize}[leftmargin=*]
  \item \textbf{Coordinate reference compatibility:} verify reference system, datum, units, axis order, and transformation path before combining or executing data.
  \item \textbf{Topology and spatial relationships:} evaluate geometry validity and required adjacency, containment, connectivity, overlap, or disjointness constraints.
  \item \textbf{Positional accuracy:} carry source accuracy or error information into downstream reasoning and reject precision claims that exceed source support.
  \item \textbf{Scale and resolution compatibility:} detect analyses that combine datasets at incompatible spatial granularity or imply unsupported detail.
  \item \textbf{Temporal validity:} record observation time, update time, effective period, and staleness constraints so that current decisions are not grounded in obsolete states.
  \item \textbf{Uncertainty and confidence:} propagate uncertainty from data, models, transformations, and derived outputs rather than treating validated values as absolute.
  \item \textbf{Jurisdiction and source authority:} determine which source has decision authority for the relevant geography, policy context, and time period.
  \item \textbf{Conflicting authoritative sources:} preserve competing evidence, compare provenance and applicability, and require adjudication when policy cannot resolve the conflict automatically.
\end{itemize}

The output of S4 is therefore not limited to valid/invalid. ANASSA may return \emph{validated}, \emph{validated-with-uncertainty}, \emph{conflicting-authoritative-evidence}, \emph{insufficient-evidence}, or \emph{rejected} states. Each state carries the evidence and validation checks that produced it.

\subsection{Component Interfaces, Execution Controls, Scalability, and Reproducibility}

To make the architecture implementable and reproducible, each inter-component transition is defined as a contract rather than an ad hoc message handoff. A contract records the requesting component, input data and provenance, spatial reference information, required tool or capability, parameters and constraints, expected output type, validation policy, timeout/retry behavior, and escalation path. The Orchestrator may only dispatch a task when its preconditions are satisfied; execution results include both the produced artifact and a machine-readable execution record.

Execution controls include tool allowlists, parameter bounds, credential and data-access constraints, timeout and retry limits, idempotency expectations where applicable, error classification, and rollback or compensating-action rules. Failed execution returns a typed failure state (for example, tool unavailable, parameter out of bounds, coordinate reference mismatch, timeout, validation rejected, or insufficient evidence) to C4 and C5 so that replanning is based on observed execution evidence rather than unconstrained model speculation.

Scalability is treated along multiple dimensions: number of participating agents, workflow depth and branching, spatial extent and resolution, data volume and update rate, distributed compute resources, and the number of organizational or jurisdictional boundaries crossed by a workflow. ANASSA does not claim a universal production scale in this paper; these dimensions define the variables that future implementations must measure.

Reproducibility is supported through a workflow provenance record containing the user objective, data identifiers and versions, models and tools invoked, workflow graph, parameters, intermediate outputs, validation results, retries, human interventions, and final disposition. This record allows a completed decision path to be reconstructed and compared across repeated runs, subject to access and confidentiality constraints.

%% ===================================================================
\balance  %% evens the columns of the page that precedes the landscape figure page
\section{Research Questions and Architectural Requirements}

The central research question motivating this study is: how can agentic GIS systems integrate spatial reasoning, workflow orchestration, validation, governance, and execution feedback within a traceable and scalable architecture? Because this paper specifies the architecture rather than reporting a domain deployment, the research questions are framed as architecture questions that can be answered through the design and subsequently tested through implementation studies.

\subsection{Research Questions}

Four architecture research questions operationalize the central question:

\begin{description}[leftmargin=*,style=unboxed]
  \item[RQ1:] What architectural capabilities are required to support spatially grounded agentic AI across distributed geospatial systems?
  \item[RQ2:] How can reasoning, planning, spatial validation, execution, and learning be coordinated within a unified spatial intelligence architecture?
  \item[RQ3:] How can authoritative spatial evidence, uncertainty, provenance, and governance constrain autonomous spatial reasoning and execution?
  \item[RQ4:] How can human expertise and decision authority be incorporated into agentic spatial workflows while preserving traceability, reproducibility, and accountability?
\end{description}

\subsection{Architectural Requirements}

The following are the design requirements for the component interfaces:

\begin{description}[leftmargin=*,style=unboxed]
  \item[DR1 --- Spatial grounding:] reasoning and execution must carry explicit coordinate reference, geometry, scale/resolution, temporal, and jurisdictional context, including sensor and platform reference frames and their transformation chain to a geodetic frame.
  \item[DR2 --- Execution-aware orchestration:] planning must account for tool availability, data dependencies, parameter constraints, runtime failures, and recovery paths.
  \item[DR3 --- Authoritative validation:] consequential spatial claims must be checked against identified evidence sources with provenance and source authority.
  \item[DR4 --- Uncertainty representation:] uncertainty and conflicting evidence must remain visible through reasoning, validation, and output generation.
  \item[DR5 --- Human decision authority:] governance must distinguish autonomous execution permitted by policy from decisions that require explicit human authorization.
  \item[DR6 --- Traceability and provenance:] each workflow must retain sufficient state to reconstruct data, models, tools, parameters, intermediate results, validations, and interventions.
  \item[DR7 --- Reproducible interfaces:] component transitions must use defined contracts and typed failure states rather than undocumented handoffs.
  \item[DR8 --- Scalable distributed execution:] the architecture must separate orchestration logic from the underlying compute, data volume, workflow depth, and organizational scale so implementations can measure and tune each independently.
\end{description}

These requirements map directly to ANASSA components: C4 addresses orchestration and execution control; C5 implements the reasoning-action loop; C7 defines governance, human authority, and auditability; C8 provides authoritative evidence, provenance, and quality metadata; C9 and C11 carry process and learning feedback; and C10 exposes validated outputs together with their decision record. This mapping is an architecture-level analysis, not an empirical performance result.

\subsection{Scope and Evaluation Positioning}

This paper defines and analyzes the ANASSA architecture; it does not report an implemented reference system, benchmark comparison, or production-scale performance result. Accordingly, the paper makes no measured claim that ANASSA improves accuracy, reduces latency, lowers cost, or outperforms existing agentic GIS systems. Such claims require controlled implementation and empirical comparison.

The present contribution instead establishes the architectural objects and measurable variables required for that evaluation: component contracts, execution success and failure stages, spatial-validation outcomes, provenance completeness, uncertainty states, human intervention points, workflow depth, data scale, and reproducibility across repeated runs. Candidate metrics include execution success rate, spatial-validation precision and recall against labeled claims, provenance-record completeness, human-escalation rate, and run-to-run output agreement. Each metric has an identifiable source in the architecture: execution success and failure stages are logged in the C4 execution records, validation outcomes in the S4 state carried to C10, provenance completeness in the C11 workflow record, and escalation rate at the C7 checkpoints. Future work should implement a reference ANASSA stack and evaluate these variables against established geospatial-agent benchmarks and domain-specific baselines.

%% ===================================================================
\clearpage
\nobalance
%% --- Physically landscape page (11 x 8.5 in) for the high-level architecture ---
\pdfpagewidth=11in \pdfpageheight=8.5in
\thispagestyle{empty}
\enlargethispage{-2.3in}
\null\vfill
\noindent\makebox[\columnwidth][l]{%
  \begin{minipage}{9.5in}
    \centering
    \Description{Block diagram of the ANASSA high-level architecture showing the eleven components C1 to C11 arranged in the Governance and Output, Operational, Cognitive Engine, and Foundation layers, with C9 and C11 drawn as vertical process flows on the left and right sides.}
    \includegraphics[width=\linewidth]{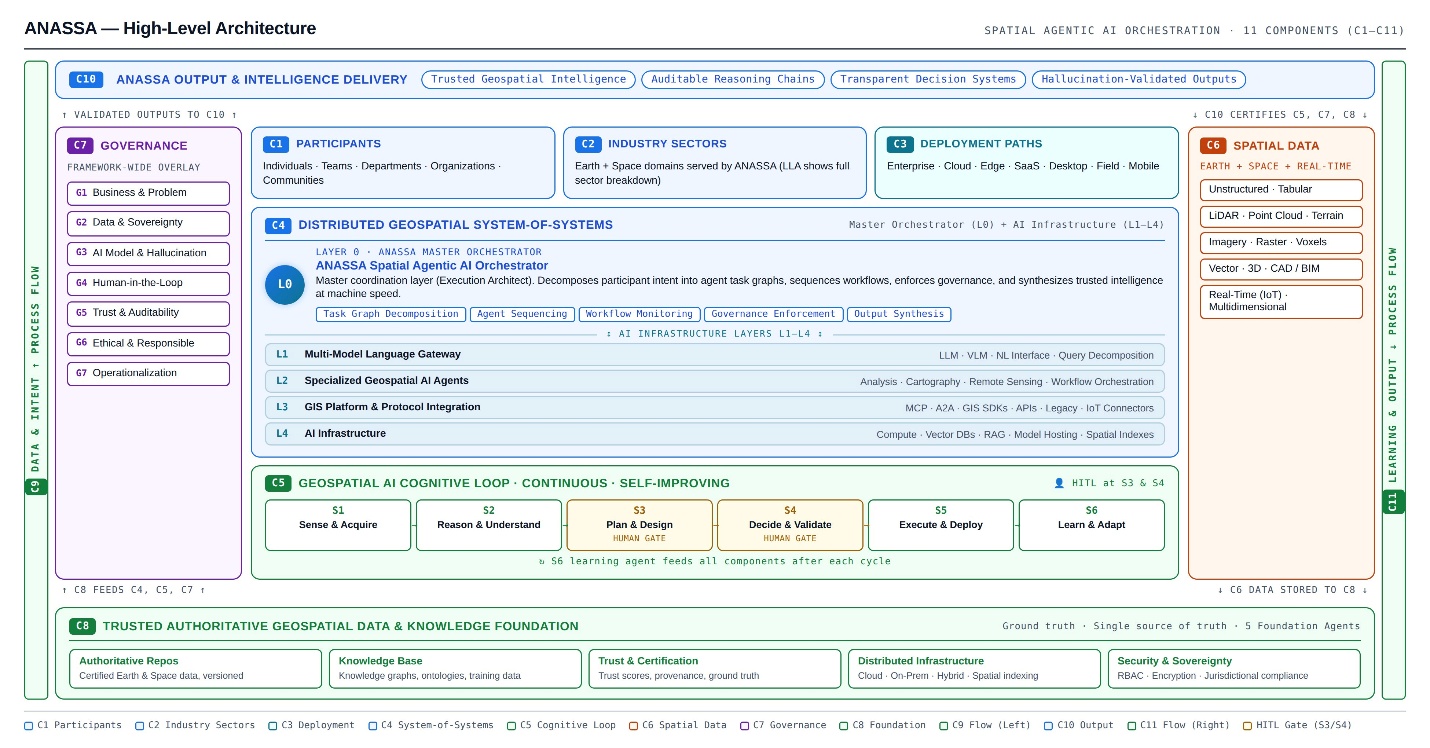}
    \captionsetup{hypcap=false}
    \captionof{figure}{ANASSA High-Level Architecture. The eleven components (C1--C11) organize across four layers: Governance and Output (C7, C10), Operational (C1--C4, C6), Cognitive Engine (C5), and Foundation (C8). C9 and C11 represent the continuous downward and upward process flows.}
    \label{fig:highlevel}
  \end{minipage}}
\vfill\null
\clearpage
\pdfpagewidth=8.5in \pdfpageheight=11in

\section{Implications and Future Research}

The ANASSA framework contributes to the emerging field of agentic GIS by addressing limitations identified in recent multi-agent and benchmarking studies and by extending prior research on spatially enabled distributed systems established by the GeoBlockchain framework. First, ANASSA integrates structured spatial reasoning with execution-aware workflow orchestration through the six-step Cognitive Loop, extending prior graph-based~\cite{bao2026spatialagent,bhattaram2025geoflow} and multi-agent~\cite{liang2026geoagenticrag,mansourian2026bridging} approaches. Second, it incorporates explicit governance, spatial validation, uncertainty, and provenance mechanisms as first-class architectural concerns rather than post-processing checks. Third, it defines cross-component contracts and typed execution states so that reasoning, tool use, validation, and recovery can be traced across heterogeneous geospatial systems. Fourth, by inheriting trust-and-accountability principles from the GeoBlockchain framework, it positions the orchestration layer itself as an auditable coordination mechanism aligned with institutional decision authority.

The study identifies six design principles for agentic GIS:

\begin{itemize}[leftmargin=*]
  \item \textbf{Integration of reasoning and execution:} Systems should tightly couple workflow generation with execution feedback rather than treating them as separate phases.
  \item \textbf{Structured intermediate representations:} Graph-based workflows and semantic abstractions improve interpretability and reliability in complex tasks.
  \item \textbf{Feedback-driven adaptation:} Robust systems require mechanisms for detecting and correcting execution failures while preserving the evidence that triggered replanning.
  \item \textbf{Spatial validation under explicit quality conditions:} Outputs should be evaluated against coordinate reference, topology, positional accuracy, resolution, temporal validity, uncertainty, jurisdiction, and source-authority constraints.
  \item \textbf{Risk-adaptive human authority:} Human intervention should be mandatory where consequence and policy require it while lower-risk workflows may operate inside approved boundaries.
  \item \textbf{Evaluation readiness and reproducibility:} Architecture-level records should make later benchmark comparison and repeated execution measurable rather than assumed.
\end{itemize}

These design principles are operationalized within ANASSA through specific architectural mechanisms. Integration of reasoning and execution is realized by the ANASSA Orchestrator (C4) in the Operational layer, which sequences reasoning and execution as one coordinated task graph and receives typed runtime feedback. Structured intermediate representations are realized through the agent task graph produced by C4 and retained in the workflow provenance record. Feedback-driven adaptation is realized by the Learning Agent at S6, subject to governance constraints on what state may be updated. Spatial validation is realized at S4 through C7/C8 checks against authoritative evidence, quality metadata, provenance, uncertainty, and source applicability. Human authority is enforced through risk- and policy-dependent checkpoints. Reproducibility is supported by the workflow provenance record. The design principles therefore correspond to identifiable architectural elements rather than unmeasured performance claims.

Future research should proceed along five connected tracks: (1) implement a reference architecture with formalized component contracts and observable execution traces; (2) benchmark reasoning, tool sequencing, failure recovery, spatial validation, and reproducibility against existing geospatial-agent baselines; (3) stress-test scalability across agent count, workflow depth, data volume, update rate, and distributed execution environments; (4) evaluate human-AI collaboration, including when policy-based automation should escalate to expert judgment; and (5) examine how uncertainty, conflicting authority, and temporal change affect decision quality. These studies should report measured outcomes separately from the architecture specification presented here.

%% ===================================================================
\section{Conclusion}

This study introduced ANASSA (Autonomous Neural Agents for Spatial Systems Architecture) as an agentic AI orchestration architecture for spatial intelligence across distributed geospatial systems. The framework responds to coordination overhead that arises when spatial intelligence systems are individually sophisticated but operationally disconnected, and to the gap between insight and controlled action that emerges when reasoning, tools, validation, governance, and data operate as separate layers of work.

The eleven components of ANASSA, from Geospatial Participants and Industry Sectors through the Distributed Geospatial System-of-Systems, Cognitive Loop, Governance and Output layer, Authoritative Geospatial Data and Knowledge Foundation, and process-flow components, together describe a coherent architecture for agentic geospatial intelligence. The architecture is grounded in prior research on spatially enabled distributed systems~\cite{papantoniou2021geoblockchain} and extends that foundation to the agentic AI domain through explicit orchestration, validation, provenance, uncertainty, and human-authority mechanisms. The paper intentionally limits its claims to architecture and evaluation readiness; implementation effectiveness remains an empirical question for subsequent studies.

ANASSA does not propose that autonomous agents replace human spatial expertise. Governance in C7 and the S3/S4 checkpoints preserve human decision authority where policy, uncertainty, conflict, or consequence requires expert judgment. The framework instead seeks to make coordination, validation, and execution traceable enough that appropriate work can be automated while human analytical capacity remains focused on interpretation, adjudication, and accountability. As with the GeoBlockchain framework that preceded it, the value proposition of ANASSA is coordination across distributed participants, agents, tools, and data through an architecture designed to be auditable and operationally accountable.

%% ===================================================================
\bibliographystyle{ACM-Reference-Format}
\bibliography{references}

\end{document}